\documentclass[letterpaper]{article} 
\usepackage{aaai2026}  
\usepackage{times}  
\usepackage{helvet}  
\usepackage{courier}  
\usepackage[hyphens]{url}  
\usepackage{graphicx} 
\usepackage{natbib}  
\usepackage{caption} 
\usepackage{algorithm}

\usepackage{newfloat}
\usepackage{listings}
\DeclareCaptionStyle{ruled}{labelfont=normalfont,labelsep=colon,strut=off} 
\floatstyle{ruled}
\newfloat{listing}{tb}{lst}{}
\floatname{listing}{Listing}
\usepackage{amsmath}
\usepackage{amsfonts}
\usepackage{IEEEtrantools}
\usepackage{booktabs}
\usepackage{makecell}
\usepackage{enumitem}
\usepackage{amssymb}
\usepackage{algpseudocode}

\title{Learning Branching Policies for MILPs with Proximal Policy Optimization}
\author {
    Abdelouahed Ben Mhamed\textsuperscript{\rm 1},
    Assia Kamal-Idrissi\textsuperscript{\rm 1},
    Amal El Fallah Seghrouchni\textsuperscript{\rm 1, \rm 2}
}
\affiliations {
    \textsuperscript{\rm 1}Ai movement - International Artificial Intelligence Center of Morocco,\\ University Mohammed VI Polytechnic, Rabat, Morocco \\
    \textsuperscript{\rm 2}Lip6, Sorbonne University, Paris, France\\
}

\begin{document}

\maketitle

\begin{abstract}
Branch-and-Bound (B\&B) is the dominant exact solution method for Mixed Integer Linear Programs (MILP), yet its exponential time complexity poses significant challenges for large-scale instances. The growing capabilities of machine learning have spurred efforts to improve B\&B by learning data-driven branching policies. However, most existing approaches rely on Imitation Learning (IL), which tends to overfit to expert demonstrations and struggles to generalize to structurally diverse or unseen instances. In this work, we propose Tree-Gate Proximal Policy Optimization (TGPPO), a novel framework that employs Proximal Policy Optimization (PPO), a Reinforcement Learning (RL) algorithm, to train a branching policy aimed at improving generalization across heterogeneous MILP instances. Our approach builds on a parameterized state space representation that dynamically captures the evolving context of the search tree. Empirical evaluations show that TGPPO often outperforms existing learning-based policies in terms of reducing the number of nodes explored and improving p-Primal-Dual Integrals (PDI), particularly in out-of-distribution instances. These results highlight the potential of RL to develop robust and adaptable branching strategies for MILP solvers.
\end{abstract}


\section{Introduction}\label{sec:intro}

Mixed Integer Linear Programming (MILP) is a foundational mathematical optimization framework for addressing complex decision-making problems characterized by both continuous and discrete variables. The inclusion of integer constraints induces non-convexity in the feasible solution space, transforming the search for optimal solutions into a combinatorial NP-hard problem. MILP has been extensively applied in various domains, including logistics and routing optimization \cite{matai2010} and energy systems planning \cite{ren2010}. This broad applicability has motivated sustained research efforts to improve solution methodologies, particularly in reducing computational complexity and accelerating solving times for large-scale real-world instances.

Branch-and-Bound (B\&B) is a canonical exact algorithm to solve discrete optimization problems, particularly integral to MILP. The method iteratively constructs a search tree where each node represents a subproblem defined by a partial assignment of integer variables. Central to B\&B are two operations: (1) branching, which partitions the feasible region of a subproblem into disjoint subsets, and (2) bounding, which computes primal and dual bounds to evaluate the subproblem's potential for improving the incumbent solution. Subproblems whose bounds exceed the current best objective value are pruned, a critical mechanism for mitigating combinatorial explosion and ensuring computational tractability \cite{land1960}.

Recent advances in Machine Learning (ML) have led to innovative methodologies to improve the efficiency of B\&B algorithms for MILP. In particular, Tree-aware branching transformers \cite{lin2022} have emerged as a paradigm for learning adaptive branching policies that generalize across MILP instances, building on earlier efforts to parameterize B\&B search trees for improved generalization \cite{zarpellon2021}. Reinforcement Learning (RL) frameworks, such as those that employ tree-structured Markov Decision Processes (MDPs), have further demonstrated the feasibility of learning branching rules from scratch without expert supervision \cite{scavuzzo2022}. Currently, advances in hierarchical representation learning have introduced transformer architectures tailored to tree-structured data, enabling the capture of long-range dependencies within B\&B search trees through multi-head attention mechanisms \cite{lin2022}. These architectures often integrate bi-directional propagation strategies to synthesize local node features with global tree context, thereby improving the fidelity of learned node representations \cite{lin2022}, \cite{zhang2025learning}. Collectively, these works underscore the potential of data-driven approaches to address the combinatorial challenges inherent in B\&B.

To address the limitations of existing imitation-based branching strategies and advance the development of generalizable policies, we propose an RL framework based on Proximal Policy Optimization (PPO). Our approach enables an agent to interact directly with the B\&B solver dynamically exploring the state-space and learning to select high value branching decisions that minimize search tree complexity.


\section{Related Work}\label{sec:related-work}

\subsection{Imitation Learning and Hybrid Frameworks}
Traditional B\&B algorithms for MILP rely on handcrafted heuristics for variable selection, which often exhibit suboptimal performance and poor scalability. Early work by \citet{he2014} demonstrated the feasibility of imitation learning to guide node selection in B\&B trees, achieving faster solve times than commercial solvers like Gurobi. However, such methods inherit biases from expert demonstrations and generalize poorly to unseen instances. Recent hybrid frameworks aim to mitigate these limitations; for example \citet{zhang2022} propose a hybrid framework that integrates imitation learning, PPO, and Monte Carlo Tree Search (MCTS) to optimize variable selection in B\&B algorithms for solving MILPs. Imitation learning bootstraps the PPO agent by mimicking strong branching heuristics, accelerating training, and reducing initial exploration inefficiencies. PPO refines the policy via stable actor-critic updates, while MCTS enhances global optimality through modified upper confidence bounds and lookahead simulations, refined via cross-entropy loss. Evaluated on benchmarks like set covering, combinatorial auctions, capacitated facility location, and maximum independent set, the method outperforms traditional heuristics and prior ML approaches in node reduction and solving times. However, limitations include potential biases from expert heuristics, high computational overhead from MCTS simulations limiting scalability, restricted evaluation to synthetic datasets questioning real-world generalization, and lingering sample inefficiency in sparse-reward settings, suggesting avenues for hybrid RL enhancements. 

\subsection{Autonomous Reinforcement Learning for Branching}\label{sec:auto-rl-branching}

To address imitation learning's limitations, later studies focused on fully autonomous RL frameworks. \citet{qu2022} introduces a Double Deep Q-Network (DDQN) based framework that leverages the demonstration data of strong branching heuristics to accelerate initial offline training, a prioritized storage mechanism to dynamically balance demonstration and self-generated data for improved policy quality, and a superior Q-network to enhance training robustness against large state-action spaces. However, limitations include the reliance on value-based DDQN, which may still suffer from overestimation biases and high variance in gradient estimation due to the expansive action spaces in MILP, potentially leading to suboptimal exploration and local optima, as evidenced by the ablation studies showing instability without the superior network; additionally, the off-policy nature and discrete action focus could limit adaptability to more dynamic or continuous branching scenarios.

\subsection{Hybrid RL with Classical Optimization}\label{sec:hybrid-rl}

Beyond pure RL, recent work integrates classical optimization techniques to enhance the performance of B\&B. \citet{prajadis2021} integrates RL with Decision Diagrams (DD) to enhance bounding mechanisms in B\&B algorithms for combinatorial optimization, specifically using Q-learning augmented by Graph Neural Networks (GNN) to learn high-quality variable orderings for approximate DD, which yield tighter bounds and reduce the size of the search tree in the maximum independent set problem compared to traditional heuristics. However, limitations arise from the computational overhead of GNN forward passes, which inflate solving times despite fewer nodes, requiring hybrid approaches with greedy heuristics and caching that can compromise bound tightness; moreover, the off-policy Q-learning framework may encounter overestimation biases, high variance in large state-action spaces, and challenges in exploration, leading to suboptimal policies or instability during training.

\subsection{Open Challenges and Our Contribution}\label{sec:challenge-contrib}

Despite notable progress in hybrid and autonomous RL for branching, three obstacles persist. 
(i) Short-horizon bias. Policies bootstrapped from strong branching can inherit short-horizon preferences for immediate bound improvements, which need not correlate with global search efficiency (e.g. total tree size). Value-based agents trained off-policy are also prone to Q-overestimation under distribution shift.
(ii) Sample inefficiency under sparse, delayed rewards. MILP solving unfolds over \(10^{3}\!-\!10^{5}\) branching decisions, where suboptimal early actions compound and informative signals (gap closure or PDI) arrive primarily at fathoming or timeout. 
(iii) Runtime overhead. Look-ahead (e.g., MCTS) and heavy GNNs can tighten bounds but increase per-node latency, stressing fixed wall-clock budgets (e.g., 3600\,s).

\paragraph{Contributions.}
We advance learning-to-branch with a fully on-policy
Tree-Gate Proximal Policy Optimization (TGPPO) architecture targeting B\&B tree size reduction.
\emph{(i) Stability without expert imprinting.}
The PPO clipped-surrogate objective mitigates off-policy estimation bias,
enabling stable training directly on solver-generated trajectories.
This avoids imprinting from hand-crafted expert traces and retains stable gradients.
\emph{(ii) Architecture for variable-arity decisions.}
We propose a permutation-equivariant Tree-Gate Transformer that
conditions attention via multiplicative gates driven by local tree statistics.
This improves credit assignment for the variable-sized candidate sets.
\emph{(iii) Throughput at scale.}
A lightweight encoder and batched rollout engine ensure
competitive wall-clock times without resorting to heavy MCTS
or GNN modules.

\noindent
\emph{Evaluation protocol.}
We use nested cross-validation with a dual criterion: (a) explored nodes for runs completed within a 1-hour budget, and (b) the PDI otherwise,
capturing both rapid completion and anytime progress.
\emph{Summary of findings.}
Across held-out benchmarks, \textsc{tgppo} surpasses prior learning-based
branchers on 72\% of test instances (fewer nodes),
while a clear, quantified gap to expert-designed heuristics
such as \textsc{relpscost} in \textsc{SCIP}\footnote{\textsc{SCIP} (Solving Constraint Integer Programs) \cite{scip} is a leading non-commercial, extensible MILP solver framework, which we use as the B\&B environment for our agent.}
still remains.

\section{Optimization and Learning Frameworks}\label{sec:optim-learn}

\subsection{Branch-and-Bound}\label{sec:bnb}

We consider a general MILP problem, formulated as:
\begin{equation}
    \begin{aligned}
        \min \quad & c^T x \\
        \text{s.t.} \quad & Ax \leq b, \\
        & x_i \in \mathbb{Z}, \quad \forall i \in \mathcal{I}, \\
        & x_j \in \mathbb{R}, \quad \forall j \in \mathcal{J}.
    \end{aligned}
    \label{eq:MILP}
\end{equation}
Here, \( x \) is the decision vector with integer variables \( x_i \) (\( i \in \mathcal{I} \)) and continuous variables \( x_j \) (\( j \in \mathcal{J} \)). The objective is defined by \( c \in \mathbb{R}^n \), and constraints by \( A \in \mathbb{R}^{m \times n} \) and \( b \in \mathbb{R}^m \). 

The B\&B algorithm solves this MILP by exploring a search tree. At each node, it solves the Linear Programming (LP) relaxation (Eq.~\eqref{eq:MILP} without integrality). If the solution $x^*$ is not integer-feasible, the algorithm \textit{branches} on a fractional variable $x_i^* \in \mathcal{I}$ by creating two subproblems with the added constraints $x_i \leq \lfloor x_i^* \rfloor$ and $x_i \geq \lceil x_i^* \rceil$. The LP solution of a subproblem provides a lower bound. A node is \textit{pruned} if its lower bound exceeds the best-known integer-feasible solution (the incumbent). This process repeats recursively until the search space is exhausted, guaranteeing optimality.

\subsection{Proximal Policy Optimization}\label{sec:ppo}

PPO \cite{schulman2017proximal} is a state-of-the-art RL algorithm that improves traditional policy gradient methods by optimizing a surrogate objective function while maintaining stable and efficient updates. PPO belongs to the class of policy-based RL algorithms, where an agent learns a parameterized policy \( \pi_{\theta}(a \mid s) \) that maps states \( s \) to a probability distribution over actions \( a \), with parameters \( \theta \) optimized to maximize the expected cumulative reward. 

A key challenge in policy optimization is to ensure a balance between policy improvement and stability. Large updates to the policy can lead to performance collapse, while overly conservative updates can slow learning. To address this, PPO introduces a clipped surrogate objective that constrains the policy update within a trust region.

The PPO objective employs an Actor-Critic architecture, where:
\begin{itemize}
    \item The \textit{actor} network $\pi_\theta(a|s)$ selects branching actions (variables) based on the current state of the solver.
    \item The \textit{critic} network $V_\phi(s)$ estimates the expected cumulative reward (the value function).
\end{itemize}

\begin{IEEEeqnarray}{c}
\mathcal{L}(\theta, \phi) = \mathbb{E}_{(s, a) \sim \pi_{\theta_{\text{old}}}} \bigg[ \min\bigg( r_{\theta}(s, a) A^{\pi_{\theta_{\text{old}}}}_{\phi}(s, a), \nonumber \\
\IEEEeqnarraymulticol{1}{c}{\text{clip}\left(r_{\theta}(s, a), 1 - \epsilon, 1 + \epsilon\right) A^{\pi_{\theta_{\text{old}}}}_{\phi}(s, a) \bigg) \bigg]} \nonumber \\
\quad - \lambda_1 \mathbb{E}_s\left[ \left(V_\phi(s) - R(s)\right)^2 \right] + \lambda_2 \mathcal{H}(\pi_\theta(\cdot|s)) \label{eq:ppo_ac}
\end{IEEEeqnarray}

where:
\begin{itemize}
    \item $r_{\theta}(s, a) = \frac{\pi_{\theta}(a \mid s)}{\pi_{\theta_{\text{old}}}(a \mid s)}$ is the probability ratio between the new and old policies.
    \item $A^{\pi_{\theta_{\text{old}}}}_{\phi}(s, a)$ is the advantage estimate, typically computed using Generalized Advantage Estimation (GAE) \cite{schulman2015high}.
    \item $R(s)$ is the empirical discounted return (the value target), used as the target for training the critic $V_{\phi}(s)$.
    \item $\mathcal{H}(\pi_\theta(\cdot|s))$ is the policy entropy that encourages exploration.
    \item $\lambda_1$, $\lambda_2$ are weighting coefficients for the critic loss and entropy bonus, respectively.
\end{itemize}

\section{Methodology}\label{sec:method}

Our methodology builds upon the feature representations introduced in prior works \cite{zarpellon2021, lin2022}, while proposing a novel RL framework based on PPO to acquire a generalized branching policy. Following \cite{zarpellon2021}, we adopt hand-crafted features. Let $\mathcal{C}_t$ be the set of candidate variables at step $t$. The state is then represented by a matrix of candidate features, $\mathbf C_t\in\mathbb{R}^{|\mathcal{C}_t| \times 25}$, and a vector of local tree state features, $Tree_t \in \mathbb{R}^{61}$. Together, these features encapsulate the evolving context of the B\&B process. In contrast to static deep neural network architectures or graph-based enhancements employed in T-BranT \cite{lin2022}, we model branching as a sequential decision-making task and train an online PPO agent to refine variable selection through direct interaction with the B\&B environment.

\subsection{Policy Architecture}

\begin{figure*}[t]
\centering
\includegraphics[width=1.0\textwidth]{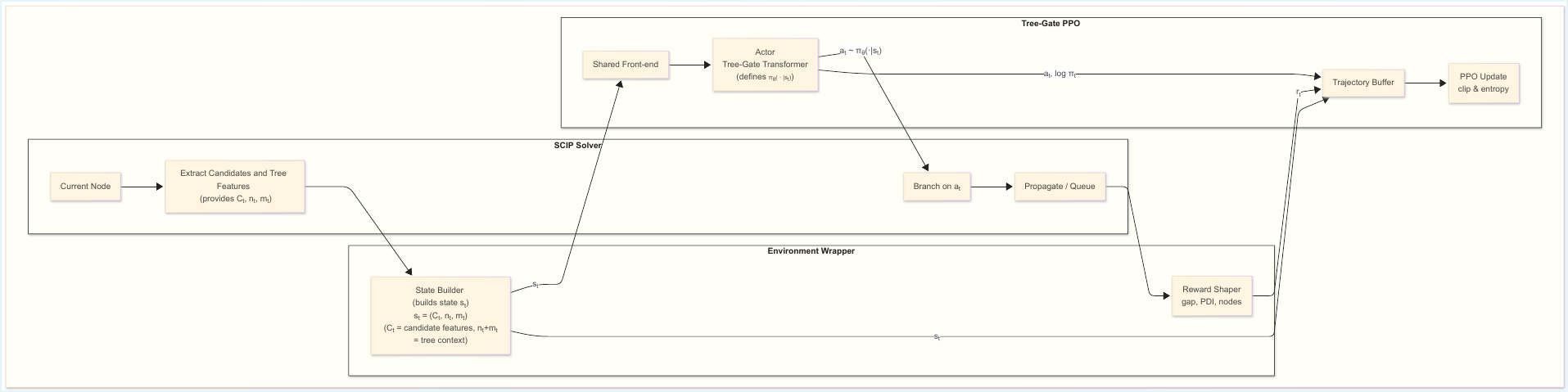} 
\caption{The Tree–Gate PPO agent interacts with SCIP through an environment wrapper: candidate features and local tree features are encoded, the actor chooses a branching variable, SCIP executes the branch, and the environment returns  rewards.  Trajectories accumulate in a buffer and are periodically used for PPO optimisation, closing the learning loop.}
\label{fig1}
\end{figure*}

Our policy follows an actor-critic design tailored to branching in MILP search trees. Each decision state comprises (i) a set of candidate branching variables with heterogeneous, instance-dependent cardinality $L$, and (ii) a tree context capturing both the current node state and local MIP features. The architecture in Figure \ref{fig1} is built around three principles: (1) permutation-equivariance over candidates; (2) contextualization of candidates by the current tree state; and (3) capacity control via a gated multi-layer reduction that adapts the effective network width to the current state. Concretely, both actor and critic share the same front-end: linear embeddings for variables and tree context, a Transformer encoder that operates over the candidate set with key-padding masks, and a light bi-directional matching module that fuses local tree information into the candidate stream. The heads then diverge: the actor produces a categorical distribution over candidates, while the critic aggregates masked candidate features into a scalar state-value.

Let $t$ index the current B\&B node. The SCIP solver
exposes three feature blocks:
\[
    \mathbf C_t\in\mathbb R^{|\mathcal{C}_t| \times d_c},\;
    \mathbf n_t\in\mathbb R^{d_n},\;
    \mathbf m_t\in\mathbb R^{d_m},
\]

where each row $\mathbf c_{t,i}$ of $\mathbf C_t$ (with
$d_c\!=\!25$) summarises candidate variable~$i$, while
$\mathbf n_t$ and $\mathbf m_t$ (with $d_n\!=\!8,\;d_m\!=\!53$)
encode the current node and MIP search tree features, respectively.

\subsubsection{Linear embeddings:}

We project every feature block into a common $d_h$-dimensional space:
\begin{equation}
\tilde{\mathbf c}_{t,i}
    =\mathbf W_c\,\mathrm{LN}(\mathbf c_{t,i}),\qquad
\tilde{\mathbf t}_t
    =\mathbf W_t\,\mathrm{LN}\!\left[\mathbf n_t;\mathbf m_t\right],
\label{eq:embeddings}
\end{equation}
with $\mathbf W_c\in\mathbb R^{d_h\times d_c}$ and
$\mathbf W_t\in\mathbb R^{d_h\times(d_n+d_m)}$ learned parameters and
$\mathrm{LN}(\cdot)$ denoting layer normalization.

\subsubsection{Candidate–tree fusion:}

By concatenating each embedded candidate with the embedded tree context and re-projecting via $\mathbf W_g \in \mathbb R^{d_h \times 2d_h}$,
\[
\mathbf z_{t,i}^{(0)}
    =\mathbf W_g\,[\tilde{\mathbf c}_{t,i};\tilde{\mathbf t}_t],
\]
we obtain the matrix
$\mathbf Z_t^{(0)}\!\in\!\mathbb R^{|\mathcal{C}_t|\times d_h}$, which is
permutation-equivariant in~$|\mathcal{C}_t|$.

\subsubsection{Self-attention encoder:}

A $N$-layer Transformer encoder maps
$\mathbf Z_t^{(0)}$ to
$\mathbf Z_t^{(N)}$ while respecting padding masks
$\mathbf M_{\text{pad}}\in\{0,1\}^{|\mathcal{C}_t|}$:
\[
\mathbf Z_t^{(N)}
    =\mathrm{Transformer}^{(N)}\bigl(\mathbf Z_t^{(0)},\mathbf
    M_{\text{pad}}\bigr).
\]

\subsubsection{Bi-directional tree–candidate matching:}
Following \citet{lin2022}, a BiMatching network refines candidate
representations via soft mutual attention:
\begin{align}
\alpha_{t,i} &=
    \mathrm{softmax}_i\bigl(
     (\mathbf W_{t1}\tilde{\mathbf t}_t)^\top \mathbf z_{t,i}^{(N)}\bigr),
\nonumber\\
\beta_{t,i} &=
    \mathrm{softmax}_i\bigl(
     (\mathbf W_{c1}\mathbf z_{t,i}^{(N)})^\top \tilde{\mathbf t}_t\bigr),
\nonumber\\
\mathbf e_t &=\textstyle\sum_{i}\alpha_{t,i}\mathbf z_{t,i}^{(N)},\quad
\mathbf d_{t,i}=\beta_{t,i}\tilde{\mathbf t}_t,
\nonumber\\[-0.25em]
\begin{split}
\mathbf r_{t,i} &= \sigma(\mathbf W_3\mathbf e_t+\mathbf W_4\mathbf d_{t,i})\!\odot\!\mathbf e_t \\
               &\quad +\bigl(1-\sigma(\mathbf W_3\mathbf e_t+\mathbf W_4\mathbf d_{t,i})\bigr)\!\odot\!\mathbf d_{t,i}
\end{split}
\label{eq:bimatch}
\end{align}
yielding refined candidate tensor
$\mathbf R_t=[\mathbf r_{t,1},\dots,\mathbf r_{t,|\mathcal{C}_t|}]^\top$.

\subsubsection{Tree-gated branching head (Actor).}
A hierarchical multi-layer perceptron of depth~$K$ reduces
each $\mathbf r_{t,i}$ to a scalar logit $\ell_{t,i}$ under the
multiplicative control of the tree vector:
\begin{align}
\mathbf g^{(k)} &=\sigma(\mathbf U_k\tilde{\mathbf t}_t),\;k=1{:}K,
\nonumber\\
\mathbf q_{t,i}^{(k)} &=
    f_k\bigl(\mathbf q_{t,i}^{(k-1)}\odot\mathbf g^{(k)}\bigr),\;
    \mathbf q_{t,i}^{(0)}=\mathbf r_{t,i},
\nonumber\\[-0.25em]
\ell_{t,i} &=\mathbf q_{t,i}^{(K)}\in\mathbb R.
\end{align}
The branching policy is the categorical distribution
$\pi_\theta(i\mid s_t)=\mathrm{softmax}_i(\ell_{t,i})$.

\subsubsection{Value head (Critic).}
A permutation-invariant state vector
$\bar{\mathbf r}_t=\frac{1}{\sum_i(1-M_{\text{pad},i})}
\sum_i (1-M_{\text{pad},i})\,\mathbf r_{t,i}$
is concatenated with $\tilde{\mathbf t}_t$ and projected through a
two-layer MLP, followed by the same tree-gated reduction, producing the
scalar value estimate $V_\phi(s_t)$.

\begin{algorithm}[t]
\caption{Tree–Gate PPO (TGPPO)}\label{alg:tgppo}
\begin{algorithmic}[1]
\Require Instance set $\mathcal D$, horizon $H$, epochs $E$
\State Initialise policy $\theta$, value $\phi$
\State $\mathcal B \gets \emptyset$ \Comment{trajectory buffer}
\ForAll{episodes}                                   \label{ln:episode-loop}
    \State Sample MILP $\mathcal I \sim \mathcal D$; reset solver
    \For{$t \gets 1$ \textbf{to} $H$}               \label{ln:time-loop}
        \State Extract features $(\mathbf C_t,\mathbf n_t,\mathbf m_t)$
        \State $a_t \sim \pi_\theta(\cdot \mid s_t)$
        \State Execute $a_t$, obtain reward $r_t$, next state $s_{t+1}$
        \State $\mathcal B \gets \mathcal B \cup (s_t,a_t,r_t,s_{t+1})$
        \If{$s_{t+1}$ is terminal \textbf{or} $|\mathcal B| = H$}
            \State Compute advantages $\hat A$ and returns via GAE
            \For{$u \gets 1$ \textbf{to} $E$}       \label{ln:epoch-loop}
                \State Sample mini-batch $\mathcal M \subset \mathcal B$
                \State Update $(\theta,\phi)$ with clipped loss Eq.~\eqref{eq:ppo_ac}
            \EndFor                                 \label{ln:epoch-loop-end}
            \State $\mathcal B \gets \emptyset$
            \State \textbf{break} \Comment{start next episode}
        \EndIf
    \EndFor                                         \label{ln:time-loop-end}
\EndFor                                            \label{ln:episode-loop-end}
\end{algorithmic}
\end{algorithm}

\subsection{Reward Signal Design}
\label{ssec:reward}

All three rewards are instance-normalized by the node, gap and PDI
statistics obtained with \textsc{SCIP}'s default branching rule
\textsc{relpscost}; this guarantees scale-robust credit assignment across
easy and hard MILPs.  Briefly:
\textbf{H1} penalizes node expansions relative to the baseline and adds a
status-dependent speed-up bonus; \textbf{H2} log-scales the penalty and
introduces a pace term plus gap/PDI shaping; \textbf{H3} adapts the
weights of those components to the problem’s difficulty, favoring gap
closure on very large instances.  The mixing coefficients were fixed
\emph{a priori} to prioritize node reduction and deliberately \emph{not}
included in the hyper-parameter search, avoiding over-tuning while keeping 
the optimization space tractable.

Since \textsc{H3} was selected as the optimal signal by our hyper-parameter search (see Section~\ref{ssec:training}), we present its core formulation here. It computes a difficulty index $d \in [0,1]$ based on the baseline node count $B$ and uses it to define adaptive weights for node efficiency ($w_{\text{nodes}}$), gap closure ($w_{\text{gap}}$), PDI ($w_{\text{pdi}}$), and pace ($w_{\text{pace}}$). The step reward is a clipped, weighted sum of these components, supplemented by a progress term $q_t$:
\begin{equation}
\label{eq:h3-step-main}
r_t^{\text{H3}} =
\mathrm{clip}_{[-1,1]}\!\Big(
w_{\text{nodes}}\,e_t
+ w_{\text{pace}}\,p_t
- w_{\text{gap}}\,g'_t
+ w_{\text{pdi}}\,d_t 
+ w_{\text{prog}}\,q_t
\Big)
\end{equation}
where $e_t$ measures log-scaled node efficiency relative to the baseline, $p_t$ is a pace-keeping term, $g'_t$ penalizes the current gap, and $d_t$ rewards PDI reduction. This is combined with a large terminal reward $R_T^{\text{H3}}$ based on the final solver status. The precise definitions of all components and the H1/H2 formulations are detailed in Appendix.

\subsection{Training Procedure}\label{sec:train}

The learning pipeline comprises two phases:
\emph{(i)} hyper–parameter selection via nested cross-validation using \textsc{Optuna}~\cite{akiba2019}, and
\emph{(ii)} final policy training with the configuration chosen in
Phase 1.
Because the public benchmark contains only $25$ MILP instances, we run
each instance under five independent SCIP seeds
($\!\text{seed}\in\{0,\dots,4\}$).
The random permutations induced by these seeds act as inexpensive,
solver-level \emph{data augmentation} that leaves the underlying
combinatorial structure unchanged, yet increases experiential diversity
five-folds.

\paragraph{Phase 1: Hyper-parameter Selection.}

\begin{description}[leftmargin=0em]
  \item[Search space.]
        The hyper-parameters and their ranges appear in
        Table~\ref{tab:search_space}.
        Exploration is performed with the TPE sampler provided by
        \textsc{Optuna}~\cite{akiba2019}.  

  \item[Nested cross-validation.]
        We adopt a two-level
        $(k_{\mathrm{outer}},k_{\mathrm{inner}})=(5,2)$ design.
        For every candidate configuration
        $\boldsymbol{\theta}\!\in\!\mathcal{H}$ we minimise, on the inner
        folds, a composite score that balances
        the number of explored nodes~$N$ (measured on runs that finish
        within the time limit) and the $\mathrm{PDI}$
        (measured on \emph{all} runs):
        \[
            \min_{\theta\in\mathcal{H}}
            \;
            \frac{1}{k_{\mathrm{in}}}
            \sum_{i=1}^{k_{\mathrm{in}}}
            \bigl(
              0.6\,\mathrm{SGM}_i(N)+0.4\,\mathrm{SGM}_i(\mathrm{PDI})
            \bigr),
        \]
        where the \emph{shifted geometric mean} (SGM) is defined as
        \[
          \mathrm{SGM}(x)=
          \exp\!\Bigl[
            \tfrac{1}{m}\sum_{j}\ln(x_j+100)
          \Bigr]-100,
        \]
        and the $+100$ shift prevents the logarithm from exploding on
        small counts.  Stratification over the
        $(\text{instance},\text{seed})$ pairs preserves the baseline
        difficulty distribution across all folds.

  \item[Pruning rule.]
        We enable Optuna’s \emph{median pruner}: a trial is stopped
        early whenever its running composite score fails to improve the
        best-so-far median across the inner folds for three consecutive
        iterations, thereby discarding unpromising settings quickly.

    \item[Selection criterion.]
        The configuration achieving the lowest outer-fold composite
        score is selected for Phase~2.
\end{description}

\paragraph{Phase 2: Final Policy Training.}
With hyper-parameters fixed, the agent is retrained on the full
$25\times5=125$ augmented problems for 500 complete episodes.  An episode corresponds to one
B\&B run up to optimality or the $3600$-second cutoff.
Seeds are reshuffled each epoch to prevent over-fitting to a fixed
permutation.

\begin{table}[t]
\centering
\small
\caption{Hyperparameter search space for Tree–Gate PPO.}
\label{tab:search_space}
\begin{tabular}{ll}
\toprule
\textbf{Parameter} & \textbf{Range / Set} \\ \midrule
Hidden size $d$ & $\{64,128,256,384\}$ \\
Transformer layers $L$ & $\{2,3,4,5,6\}$ \\
Attention heads $h$ & $\{2,4,8\}$ \\
Dropout $\delta$ & $[0,0.3]$ \\
Actor LR $\alpha_{\mathrm{act}}$ & $[10^{-6},3\!\times\!10^{-4}]$ \\
Critic LR $\alpha_{\mathrm{crt}}$ & $[10^{-6},3\!\times\!10^{-4}]$ \\
PPO clip $\epsilon$ & $[0.05,0.3]$ \\
Entropy weight $\beta$ & $[10^{-5},10^{-2}]$ \\
GAE $\gamma$ & $[0.92,0.999]$ \\
GAE $\lambda$ & $[0.8,0.99]$ \\
Minibatch size $b$ & $\{32,64,128,256,512\}$ \\
Epochs $E$ & $\{1,2,3,4,5,6\}$ \\
Reward & $\{H1, H2, H3\}$\\ \bottomrule
\end{tabular}
\end{table}


\section{Experiments}
\subsection{Datasets} \label{sec:datesets}

Following \citet{lin2022}, we train and evaluate on the exact
same curated benchmark drawn from the public MILP libraries
MIPLIB~3/2010/2017 and CORAL\footnote{All instances are distributed
under the CC BY–SA~4.0 licence.}.  
\citet{lin2022} ran \textsc{SCIP}\,6.0.1 with the default
\textsc{relpscost} rule on every publicly available instance and kept
only those that (i) are feasible and bounded, (ii) achieve a finite
primal gap within a $7200$\,s limit, and (iii) contain no more than
$200\,000$ variables, discarding the rest.
The resulting collection comprises 25 training instances and
66 testing instances as shown in table \ref{tab:instances}; the test set is further split into
33 easy and 33 hard problems according to whether all
policies (except random) solve them within $3600$\,s.
Using the same dataset ensures strict comparability with prior
learning-based branching work while covering a heterogeneous mix of
set–covering, facility-location, auction, and scheduling models.

\begin{table}[t]
\centering
\small
\setlength{\tabcolsep}{3pt}        
\caption{List of training and test instances selected from MIPLIB and CORAL.}
\begin{tabular}{@{}p{0.94\linewidth}@{}}
\toprule
\textbf{Train (25):} 30n20b8, air04, air05, cod105, comp21-2idx, dcmulti, eil33-2, istanbul-no-cutoff, l152lav, lseu, misc03, neos20, neos21, neos-476283, neos648910, pp08aCUTS, rmatr100-p10, rmatr100-p5, rmatr200-p5, roi5alpha10n8, sp150$\times$300d, stein27, supportcase7, swath1, vpm2 \\[4pt]
\midrule
\textbf{Test (66):} aflow40b, app1-2, atlanta-ip, bab5, bc1, bell3a, bell5, biella1, binkar10\_1, blend2, dano3\_5, fast0507, harp2, map10, map16715-04, map18, map20, mik-250-20-75-4, mine-166-5, misc07, msc98-ip, mspp16, n2seq36q, n3seq24, neos11, neos12, neos-1200887, neos-1215259, neos13, neos18, neos-4722843-widden, neos-4738912-atrato, neos-480878, neos-504674, neos-504815, neos-512201, neos-584851, neos-603073, neos-612125, neos-612162, neos-662469, neos-686190, neos-801834, neos-803219, neos-807639, neos-820879, neos-829552, neos-839859, neos-892255, neos-950242, ns1208400, ns1830653, nu25-pr12, nw04, opm2-z7-s2, p0201, pg, pigeon-10, pp08a, rail507, roll3000, rout, satellites1-25, seymour1, sp98ir, unitcal\_7 \\
\bottomrule
\end{tabular}
\label{tab:instances}
\end{table}

\subsection{Settings}
\label{ssec:settings}

Our experimental framework utilizes SCIP 6.0.1 as the underlying MILP solver, interfaced via a customized version of \textsc{PyScipopt} \cite{maher2016} to access the solver’s internal state and decision points required for training the PPO policy. This customization builds on prior work by \citet{zarpellon2021}, which demonstrated the feasibility of integrating ML with B\&B solvers through parameterized search trees. To isolate the impact of branching variable selection (BVS) and align with our proof-of-concept study design, we adopt a streamlined solver configuration following established BVS benchmarking practices \cite{Linderoth1999}. Specifically, we disable primal heuristics and provide the known optimal solution value as an initial cutoff, ensuring the solver focuses exclusively on tree-search efficiency rather than primal bound discovery \cite{Gamrath2017}. Although this represents a simplification of a full solver run, this ``expert'' setting is a standard benchmarking practice to isolate the performance of branching rules, as done in \cite{lin2022, zarpellon2021}. We analyze the impact of this choice in our limitations (\ref{sec:conclusion}). A one-hour time limit per instance balances computational feasibility with solution quality assessment. 

\begin{figure*}[t]
  \centering
  \includegraphics[width=0.49\textwidth]{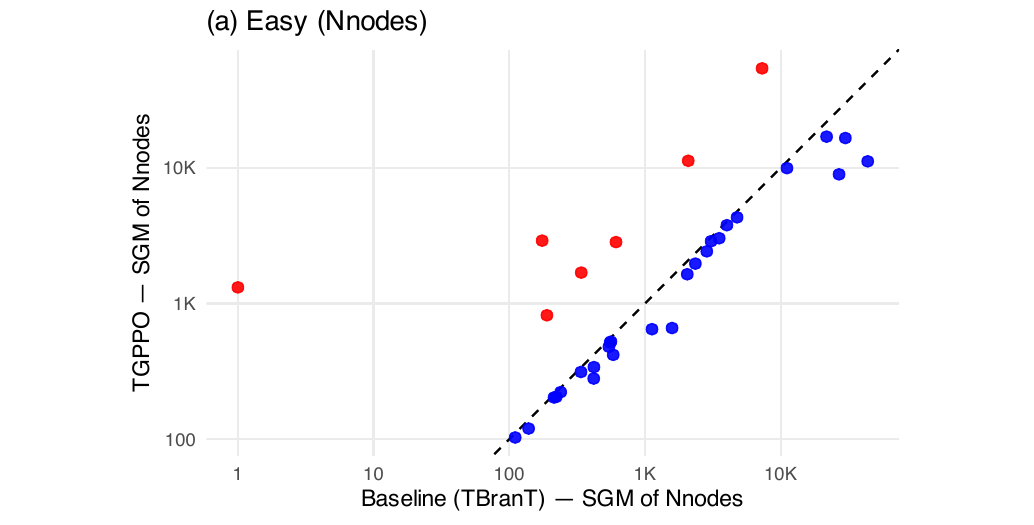}\hfill
  \includegraphics[width=0.49\textwidth]{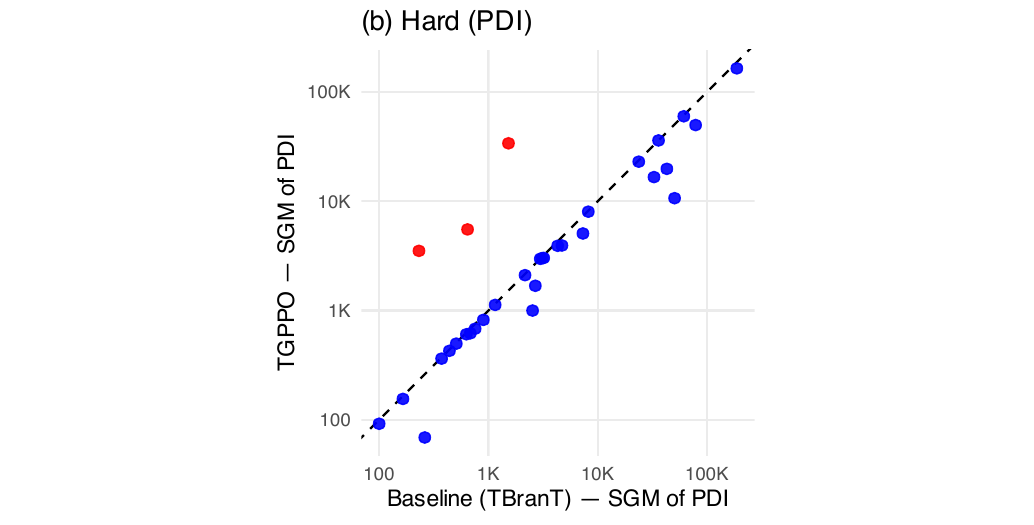}
  \caption{Head-to-head scatter of \textsc{tgppo} vs.\ \textsc{tbrant} (log–log): points below the dashed diagonal ($y<x$) indicate \textsc{tgppo} wins; (a) comparison of the easy instances using Nnodes, (b) comparison of the hard instances using PDI}
  \label{fig:head2head}
\end{figure*}

\subsection{Training}
\label{ssec:training}

After the nested cross-validation described in
\ref{sec:train}, the best hyper-parameter configuration was found to
be:
\textit{actor\_lr} $=2.4\times10^{-4}$,
\textit{critic\_lr} $=1.2\times10^{-4}$,
hidden size $d_{\mathrm{model}}=256$,
$L=5$ transformer layers,
$H=8$ attention heads,
dropout $0.05$,
PPO clip coefficient $0.16$,
entropy bonus $3.0\times10^{-3}$,
$\gamma=0.97$,
$\lambda_{\mathrm{GAE}}=0.92$,
mini-batch size~256,
three optimization epochs per update, and the
difficulty–adaptive reward \textsc{H3}.
We then retrained a fresh policy with these
values on the full training dataset.

Training proceeds for 500 complete episodes.
Each episode consists of a full B\&B run
(initialized with a new seed) and is followed by three
gradient-descent epochs over the collected trajectories.
Both actor and critic are optimized with AdamW
($\beta_1=0.9,\ \beta_2=0.999$);
the PPO objective includes the above clip
coefficient and an entropy regularizer of
$3.0\!\times\!10^{-3}$.
All experiments are implemented in PyTorch \cite{Paszke2019}
and executed on 40-core CPU nodes.

\subsection{Results and Analysis}\label{ssec:results}

\paragraph{Protocol and metrics.}
Each test instance (\ref{sec:datesets}) is evaluated under five independent seeds (0--4), yielding \(33\times5=165\) runs for the \emph{easy} subset and the same for the \emph{hard}.
We record (i) explored nodes (\textsc{Nnodes}) and (ii) (\textsc{PDI}).
On easy instances we compare \textsc{Nnodes}; on hard instances, we compare \textsc{PDI} (lower is better).

\paragraph{Aggregates.}
We report two complementary views: (1) an \emph{overall} shifted geometric mean
\[
  \mathrm{SGM}(x_{1{:}m})=\exp\!\Bigl[\tfrac1m \sum_{j=1}^m \ln(S+x_j)\Bigr]-S,
\]
computed across all instance–seed pairs (with $S{=}100$ for \textsc{Nnodes} and $S{=}0$ for \textsc{PDI}); and
(2) a \emph{per-instance} comparison: first compute the SGM over the five seeds of an instance, then compare \textsc{tgppo} head-to-head against each baseline (a win is counted if \textsc{tgppo}'s SGM is strictly smaller).

\paragraph{Head-to-head dominance.}
Table~\ref{tab:winrate} shows the fraction of instances on which \textsc{tgppo} dominates each baseline (SGM over seeds).
\textsc{tgppo} surpasses the prior state-of-the-art learner \textsc{tbrant} on $78.8\%$ of instances by \textsc{Nnodes} and $90.6\%$ by \textsc{PDI}; see Fig.~\ref{fig:head2head}, where points below the dashed diagonal indicate \textsc{tgppo} wins.
It also dominates other learning branchers (\textsc{brant}, \textsc{ltbrant}, \textsc{tree}) on roughly $73\%$ by \textsc{Nnodes} and $\approx\!69$--$72\%$ by \textsc{PDI}.
Among the classical rules, \textsc{pscost} is most competitive (wins $66.7\%$ / $78.1\%$ for \textsc{Nnodes}/\textsc{PDI}), while \textsc{relpscost} is harder to beat in nodes ($18.2\%$) but is still outperformed in \textsc{PDI} nearly half the time ($46.9\%$).
These outcomes indicate consistent per-instance superiority of \textsc{tgppo}, even though some instances still favor hand-crafted rules.

\begin{table}[t]
\centering
\footnotesize
\setlength{\tabcolsep}{6pt}
\caption{Per–instance dominance of \textsc{tgppo}. Entries are the \% of test instances for which \textsc{tgppo} improves over the baseline (SGM over five seeds).}
\label{tab:winrate}
\begin{tabular}{lcc}
\toprule
\textbf{Baseline} & \textbf{\% win (Nnodes)} & \textbf{\% win (PDI)} \\
\midrule
\textsc{tbrant} \cite{lin2022}      & 78.8 & 90.62\\
\textsc{brant} \cite{lin2022}       & 72.7 & 71.87\\
\textsc{ltbrant} \cite{lin2022}     & 72.7 & 71.87\\
\textsc{tree} \cite{zarpellon2021}& 72.7 & 68.75\\
\midrule
\textsc{random}      & 90.9 & 93.75\\ 
\textsc{pscost}      & 66.7 & 78.12\\
\textsc{relpscost}   & 18.2 & 46.87\\
\bottomrule
\end{tabular}
\end{table}

\begin{table*}[t]
\centering
\footnotesize
\setlength{\tabcolsep}{4pt}
\caption{Compact omnibus and post-hoc: Friedman test on within-instance ranks and Nemenyi $p$ (single-step) for \textsc{tgppo} vs.\ others. Bold $p{<}.05$.}
\label{tab:stats-compact}
\begin{tabular}{lcccccc}
\toprule
\textbf{Subset} & \(\chi^2\) (df) & \(\mathbf{p}\) & \textsc{brant} & \textsc{ltbrant} & \textsc{tbrant} & \textsc{tree} \\
\midrule
Easy (\textsc{Nnodes}) & 22.215 (4) & 0.0001816 & \textbf{0.0064} & \textbf{0.0027} & 0.5000 & \textbf{0.0011} \\
Hard (\textsc{PDI})    & 18.05 (4)  & 0.001207  & \textbf{0.0026} & \textbf{0.0105} & \textbf{0.0046} & 0.0557 \\
\bottomrule
\end{tabular}
\end{table*}

\paragraph{Overall comparisons (easy: \textsc{Nnodes}; hard: \textsc{PDI}).}
We analyze within-instance ranks with a Friedman omnibus test \cite{friedman1937use} and Nemenyi post-hoc \cite{nemenyi1963distribution} (single-step adjusted $p$). The easy subset shows significant differences across the five learning policies (\(\chi^2{=}22.215\), $df{=}4$, $p{=}1.82{\times}10^{-4}$), with \textsc{tgppo} significantly better ranked than \textsc{brant}, \textsc{ltbrant}, and \textsc{tree}, and statistically tied with \textsc{tbrant}.
On the hard subset, ranks by \textsc{PDI} also differ significantly (\(\chi^2{=}18.05\), $df{=}4$, $p{=}0.00121$); \textsc{tgppo} is significantly better ranked than \textsc{brant}, \textsc{ltbrant}, and \textsc{tbrant}, and marginal vs.\ \textsc{tree}.
For directionality on raw values, one-sided paired Wilcoxon tests favor \textsc{tgppo} on easy instances (wins on $24$--$26$ of $33$; median node reductions $80$--$142$; Holm-adjusted $p\in[0.09,\,0.11]$) and on hard instances (wins on $22$--$29$ of $33$; median \textsc{PDI} reductions $82$--$111$; adjusted $p\le 0.0314$).

\paragraph{Practical takeaway.}
\textsc{tgppo} reduces search effort on easy instances and, critically, yields markedly better \textsc{PDI} on hard instances where optimality is not reached within one hour, a regime of substantial practical interest.


\section{Conclusion and Perspectives}
\label{sec:conclusion}

This paper introduced \textsc{tgppo}, an on-policy RL framework that learns branching policies for MILPs directly from solver interactions. The approach combines (i) a permutation–equivariant transformer encoder over candidate variables, (ii) multiplicative tree-gates that condition decisions on local node and tree statistics, and (iii) a PPO training loop with instance-normalized rewards and a nested cross-validation protocol for robust model selection. Empirically, \textsc{tgppo} consistently improves search efficiency over prior learning-based branchers. On easy instances it reduces the number of explored nodes, while on hard instances (where runs may time out) it delivers markedly lower PDIs. Together, these results indicate that on-policy RL can be a competitive and stable alternative to imitation-driven methods for learning-to-branch, narrowing, though not erasing, the gap to strong, hand-crafted rules such as \textsc{relpscost}.

\paragraph{Limitations and threats to validity.}
Our experimental protocol focuses on branching-variable selection under a streamlined solver configuration to isolate branching effects; this leaves end-to-end interactions with primal heuristics and other solver modules for future study. The training set is necessarily small at the instance level and curated for comparability with prior work, which may limit coverage of industrial distributions. Although our instance-normalized rewards improve stability, reward shaping remains a sensitive design choice. Finally, while our rollout engine provides solid CPU throughput, full system-level benchmarking on diverse hardware and wall-clock metrics (including memory footprint) is still required for operational deployment.

\paragraph{Perspectives.}
We see several promising directions to extend this work:
\begin{enumerate}[leftmargin=*,itemsep=2pt,topsep=2pt]
    \item \textbf{End-to-end solver integration.} Re-enable primal heuristics, cuts, and default parameterizations to evaluate the net impact on wall-clock time, optimality gaps, and anytime behavior; formulate training as a multi-objective problem balancing nodes, PDI, and time.
    \item \textbf{Representation learning.} Enrich the encoder with bipartite variable–constraint message passing or lightweight decision-diagram features; add memory over partial subtrees to better capture long-range dependencies without heavy graph stacks.
    \item \textbf{Data- and compute-efficiency.} Explore offline pretraining on solver logs followed by on-policy fine-tuning (Behavior Cloning $\rightarrow$ PPO), and investigate advantage normalization, return decomposition, or truncated credit assignment tailored to deep search trees.
    \item \textbf{Reward design and objectives.} Systematically study difficulty-adaptive rewards and risk-sensitive objectives (e.g., CVaR/quantile variants) to trade off mean performance and tail robustness on hard instances.
\end{enumerate}

In sum, \textsc{tgppo} advances learning-to-branch by coupling stable on-policy optimization with a tree-aware architecture that respects the symmetries of candidate sets and the context of the search. We expect that integrating richer representations, end-to-end objectives, and deployment-conscious systems design will further tighten the gap to expert heuristics and make RL-based branching practical for large, heterogeneous MILPs.

\bibliography{aaai2026}

\appendix

\begingroup
\setlength{\abovedisplayskip}{6pt}
\setlength{\belowdisplayskip}{6pt}
\setlength{\abovedisplayshortskip}{4pt}
\setlength{\belowdisplayshortskip}{4pt}
\small

\section{Reward Formulations}
\label{app:rewards}

\paragraph*{Notation.}
Let $B$ denote the baseline node count from \textsc{SCIP}’s
\textsc{relpscost} (per instance), $n_t$ the cumulative nodes explored
by our policy up to step $t$, and $\Delta n_t \!=\! n_t{-}n_{t-1}$.
Let $\Delta_{\text{gap}}^{t}$ be the relative optimality gap at $t$,
$\text{PDI}^{t}$ the primal–dual integral, and
$\tau_t \!=\! t/T_{\max}\!\in[0,1]$ the normalized wall time.
At termination ($t{=}T$) the solver status $\text{stat}_T\in
\{\textsc{optimal},\textsc{timelimit},\textsc{infeasible},\textsc{unbounded}\}$.
We use the numerically safe mappings
$\mathrm{tanh}_s(x)=\tanh(sx)$ and
$\mathrm{ratio}_c(a,b)=\min(a/\max\{b,10^{-12}\},c)$.

\paragraph*{H1: Baseline-Normalised Node Efficiency.}
Step penalty and terminal bonus:
\begin{align}
r_t^{\text{H1}} &=
-\mathrm{tanh}_\alpha\!\left(
      \frac{\Delta n_t}{0.02\,B + 1}
     \right),\quad \alpha{=}1,
\label{eq:h1-step}\\[-2pt]
R_{\!T}^{\text{H1}} &=
\begin{cases}
  1 + 2s, & \textsc{optimal},\\
  0.5 + 1.5s, & \textsc{infeasible}/\textsc{unbounded},\\
  0.2s + 0.6\,g_T + 0.2\,d_T, & \textsc{timelimit},\\
  0.2s, & \text{otherwise},
\end{cases}
\label{eq:h1-term}
\end{align}
where $s=\mathrm{ratio}_c(B,n_T;3)$,
$g_T=\mathrm{tanh}_1(\Delta_{\text{gap}}^{0}-\Delta_{\text{gap}}^{T})$,
and
$d_T=\mathrm{tanh}_1\!\big((\text{PDI}^{0}-\text{PDI}^{T})/\text{PDI}^{0}\big)$.

\paragraph*{H2: Log-Scaled Efficiency with Pace Shaping.}
With $\beta{=}1.5$, $\rho{=}0.7$:

\begin{align}
e_t &= \tanh_\beta\!\left( 1-\frac{\log(1+n_t)}{\log(1+B)} \right),\\[2pt]
p_t &= \tanh_\beta\!\left( \frac{B\tau_t^{\rho}-n_t}{B\tau_t^{\rho}+1} \right).
\end{align}

\begin{align}
g_t &=
\begin{cases}
\tanh_1\!\bigl(\frac{\Delta_{\!{\rm gap}}^{t-1}-\Delta_{\!{\rm gap}}^{t}}
          {|\Delta_{\!{\rm gap}}^{t-1}|+10^{-9}}\bigr), & t>0,\\
0,& t=0,
\end{cases}\\[4pt]
d_t &= \tanh_1\!\left(\frac{{\rm PDI}^{t-1}-{\rm PDI}^t}{{\rm PDI}^0}\right).
\end{align}

\begin{align}
r_t^{\text{H2}} &=
\operatorname{clip}_{[-1,1]}
\!\left(0.5e_t + 0.2p_t + 0.2g_t + 0.1d_t\right).
\end{align}

\begin{align}
R_T^{\text{H2}} &=
\begin{cases}
1+2.5s, & \textsc{optimal},\\
0.7+2s, & \textsc{infeasible}/\textsc{unbounded},\\
0.4s + 0.4g_T + 0.2d_T, & \textsc{timelimit},\\
0.3s, & \text{otherwise}.
\end{cases}
\end{align}

\paragraph*{H3: Difficulty-Adaptive Reward.}
Define a difficulty index

\[
d=\sigma\!\left(
    \frac{\log(1+B)-\log 2}{\log(1+10^{6})-\log 2}
\right)\!\in[0,1].
\]

\begin{align}
w_{\text{nodes}} &= 0.55(1-d)+0.25d,\\
w_{\text{gap}}   &= 0.10(1-d)+0.30d,\\
w_{\text{pdi}}   &= 0.05(1-d)+0.20d,\\
w_{\text{pace}}  &= 0.15(1-d)+0.10d,\\
w_{\text{prog}}  &= 0.15.
\end{align}

\begin{align}
r_t^{\text{H3}}
&= \operatorname{clip}_{[-1,1]}\!\Big(
    w_{\text{nodes}}\,e_t
  + w_{\text{pace}}\,p_t
  - w_{\text{gap}}\,
    \tanh_{0.5}\!\Bigl(\notag\\
&\qquad\qquad
      \frac{\Delta_{\text{gap}}^{t}}
           {\Delta_{\text{gap}}^{0}+10^{-9}}
    \Bigr)
  + w_{\text{pdi}}\,d_t
  + w_{\text{prog}}\,q_t
\Big).
\end{align}

\begin{align}
R_T^{\text{H3}} &=
\begin{cases}
1 + 3s, & \textsc{optimal},\\
0.8 + 2s, & \textsc{infeasible}/\textsc{unbounded},\\
0.5s + 0.3\,g_T + 0.2\,d_T, & \textsc{timelimit},\\
0.3s, & \text{otherwise}.
\end{cases}
\end{align}
\paragraph*{Weight choice.}
The component weights in H2/H3 were fixed a priori to prioritize
shrinking the search tree (node efficiency) while preserving anytime
progress (gap/PDI). We deliberately \emph{did not} tune these weights
during hyper-parameter search to avoid expanding the search space and to
keep the evaluation reproducible.

\section{Additional Results}
\label{app:results}

\begin{table*}[t]
\centering
\scriptsize
\setlength{\tabcolsep}{3pt}
\caption{Number of nodes explored by each branching policy. Bold figures indicate that \textsc{tgppo} outperforms
\textsc{tbrant} on the same instance.}
\label{tab:nodes}
\begin{tabular}{lrrrrrrrr}
\toprule
\textbf{Instance} &
\textbf{brant} &
\textbf{lbrant} &
\textbf{pscost} &
\textbf{random} &
\textbf{relpscost} &
\textbf{tbrant} &
\textbf{tgppo} &
\textbf{tree} \\
\midrule
ALL TEST     & 1602.91  & 1484.65  & 1822.01  &  4787.13  &   731.37  &  1318.33 & 1588.72 &  1671.27 \\
\midrule
aflow40b      & 21357.41 & 18275.65 & 19828.14 & 274377.17 &  9331.03 & 21795.26 & \textbf{16971.64} & 21745.75 \\
bc1           &  4875.32 &  4391.09 &  4269.85 &   7420.14 &  2565.54 &  4771.82 & \textbf{4315.00} &  4988.66 \\
bell3a        &  2235.47 &  1964.98 &  1502.46 &   2033.78 &  2025.22 &  2050.18 & \textbf{1643.00} &  2029.09 \\
bell5         &   355.11 &   424.39 &   566.32 &    359.41 &   362.81 &   419.75 & \textbf{280.04}  &   392.04 \\
binkar10\_1   &  3375.36 &  3274.07 &  3242.54 &   5090.92 &  2489.30 &  3072.18 & \textbf{2879.40} &  4273.29 \\
blend2        &   515.78 &   525.50 &   621.53 &   1031.12 &   172.94 &   420.43 & \textbf{340.00}  &   356.27 \\
dano3\_5      &   219.21 &   229.95 &   239.48 &    435.13 &    54.47 &   138.81 & \textbf{119.92}  &   248.46 \\
map20         &  2504.94 &   341.74 &   800.78 &   6389.80 &   227.48 &   337.59 & \textbf{313.00}  &  4611.05 \\
misc07        & 26380.91 & 25763.24 & 17822.84 &  39116.32 & 29191.05 & 26930.40 & \textbf{8952.64} & 28352.56 \\
n2seq36q      &  1354.12 &   800.41 &  1444.75 &    729.64 &  1077.22 &  1126.69 & \textbf{647.00}  &   740.33 \\
neos-1200887  & 56066.21 & 61136.40 & 53279.03 & 995155.72 & 250472.28 & 43785.80 & 111624.42 & 48046.76 \\
neos-1215259  & 28570.73 & 26758.50 &  7896.79 &  73239.39 &   893.60 &  2358.62 & \textbf{1968.15} &  2365.55 \\
neos-504674   & 15322.44 & 18949.81 & 11840.86 &  84011.86 &  7740.03 &  7295.20 & 54057.68 & 18558.15 \\
neos-504815   &  3817.37 &  4143.30 &  3751.79 &  16947.46 &  2527.60 &  3523.58 & \textbf{3028.15} &  3683.93 \\
neos-512201   &  6406.57 &  4388.47 &  2816.04 &  13815.58 &  1508.07 &  2084.19 & 11264.13 &  3102.75 \\
neos-584851   &  1488.80 &  1669.34 &  2011.74 &   2718.10 &   231.42 &   583.53 & \textbf{418.62}  &  2327.39 \\
neos-612125   &   122.74 &   122.70 &   178.99 &    547.69 &    69.25 &   110.57 & \textbf{103.00}  &   115.91 \\
neos-612162   &   241.40 &   247.21 &   290.31 &    911.83 &   138.31 &   221.57 & \textbf{205.00}  &   197.78 \\
neos-801834   &   605.95 &   716.27 &  1339.37 &   1282.85 &   230.62 &   553.97 & \textbf{517.00}  &   670.64 \\
neos-803219   & 33220.80 & 26894.64 & 30731.61 & 112534.17 & 12394.67 & 29982.03 & \textbf{16575.41} & 26737.87 \\
neos-807639   & 12315.86 & 10683.13 &  5190.83 &  20090.11 &  3880.27 & 11128.42 & \textbf{9945.08} & 14668.51 \\
neos-820879   &   352.41 &   398.22 &   927.56 &   3838.14 &   125.90 &   339.23 & 1690.00 &   408.96 \\
neos-892255   &   878.62 &   746.42 &   736.23 &   1048.06 &   636.76 &   611.80 & 2835.74 &  1085.85 \\
neos-950242   &  3392.59 &  3602.97 &  4728.49 &  29949.69 &  2618.27 &  2854.37 & \textbf{2422.81} &  2541.50 \\
ns1208400     &   204.94 &   281.61 &  1187.38 &   1640.88 &    58.75 &   174.73 & 906.98 &   267.86 \\
nu25-pr12     &   217.14 &   264.12 &   342.47 &   1658.41 &    21.39 &   189.39 & 819.00 &   204.81 \\
nw04          &     1.00 &     1.00 &    71.08 &    135.64 &     1.00 &     1.00 & 1315.00 &     1.00 \\
p0201         &   161.82 &   186.23 &   200.86 &    219.77 &    18.51 &   213.65 & \textbf{203.00} &   232.61 \\
pg            &   553.12 &   575.58 &   565.24 &    652.29 &   134.59 &   561.26 & \textbf{523.99} &   570.56 \\
pp08a         &   208.94 &   214.49 &   254.26 &    304.71 &    45.04 &   239.32 & \textbf{222.51} &   221.77 \\
satellites1-25&   917.57 &   700.71 &  3050.29 &   5051.30 &   530.28 &  1583.85 & \textbf{659.70} &  1625.83 \\
sp98ir        &  3096.86 &  4080.53 &  5663.54 &  39208.94 &  1225.49 &  4016.84 & \textbf{3776.87} &  4425.60 \\
unitcal\_7    &  621.67 &  590.95 &  656.11 &  2429.35 &  197.43 &  542.50 & \textbf{480.00} &  1433.84 \\
\bottomrule
\end{tabular}
\end{table*}

\begin{table*}[t]
\centering
\scriptsize
\setlength{\tabcolsep}{3pt}
\caption{PDI of each policy.
Bold figures indicate that \textsc{tgppo} outperforms
\textsc{tbrant} on the same instance.}
\label{tab:pdi}
\begin{tabular}{lrrrrrrrr}
\toprule
\textbf{Instance} &
\textbf{brant} &
\textbf{lbrant} &
\textbf{pscost} &
\textbf{random} &
\textbf{relpscost} &
\textbf{tbrant} &
\textbf{tgppo} &
\textbf{tree} \\
\midrule
ALL TEST          & 3347.67  & 2776.06  & 3879.65  & 10404.71  & 1981.85 & 2922.72 & 2945.33 & 2939.30 \\
\midrule
app1-2            & 48065.58 & 64566.58 & 17571.65 & 74204.23 & 48510.64 & 50486.20 & \textbf{10662.62} & 68145.71 \\
atlanta-ip        & 27341.37 & 25261.71 & 33241.90 & 16611.90 & 16161.90 & 23019.96 & 25162.92 & 25152.92 \\
bab5              &  2950.77 &  3300.29 &  3649.85 &  3515.74 &  2857.45 &  3104.14 & \textbf{3017.28} &  3161.85 \\
bell4             &   389.60 &   359.27 &   415.42 &   751.32 &   338.24 &   373.05 & \textbf{336.38} &   386.70 \\
fast0507          &   759.61 &   759.04 &  1008.48 &  4282.23 &   853.35 &   756.44 & \textbf{682.07} &   750.39 \\
harp2             &  1068.41 &    0.00  &    75.20 &   253.58 &    26.20 &   165.24 & \textbf{155.52} &   106.20 \\
map10             & 33678.51 &  8603.33 & 13714.21 & 33711.26 &  6860.22 &  8202.81 & \textbf{8026.37} & 29696.21 \\
map16715-04       &153711.00 & 68731.63 &139907.50 &194165.73 &109656.59 & 61628.88 & \textbf{59857.30} &183932.06 \\
map18             &   249.33 &  1207.50 &  2670.84 & 14625.19 &  1734.81 &  1153.24 & \textbf{1126.14} &  3213.67 \\
mik-250-20-75-4   &   342.36 &   374.06 &   102.46 &  3788.17 &   107.03 &   644.41 &  5522.93 &   256.99 \\
mine-166-5        &   589.38 &   621.70 &  1418.08 &  9936.65 &   592.82 &   628.77 & \textbf{608.52} &   604.81 \\
msc98-ip          &  3063.46 &  2561.40 &  3048.09 & 30005.12 &  2759.71 &  2982.23 & \textbf{2971.23} &  2171.01 \\
mspp16            & 31422.96 & 92314.93 & 34335.58 & 38343.58 & 27227.17 & 32715.92 & \textbf{1662.65} & 35488.06 \\
n2seq24           &  7441.40 &  7705.33 &  7222.47 &  6500.93 &  7383.59 &  7321.92 & \textbf{5070.49} &  7881.80 \\
neos-4722843-widden& 82183.73& 87286.31 & 73739.80 &120234.25 & 75606.13 & 78598.83 & \textbf{64733.27} & 71320.47 \\
neos-4738912-atrato&  6130.94&  4726.75 &  3786.63 &  3270.19 &  1875.18 &  2531.34 & \textbf{  999.91} &  1975.43 \\
neos-480878       &   144.58 &   144.85 &   144.75 &  3461.05 &    68.94 &   231.52 & 3516.70 &   111.06 \\
neos-603073       & 15230.76 &  6017.58 & 10356.21 & 19456.74 & 34982.95 &  4302.37 & \textbf{3911.20} & 18902.46 \\
neos-662469       &   544.29 &   516.65 &   485.98 &   706.49 &   483.42 &   508.64 & \textbf{498.46} &   514.40 \\
neos-686190       &   643.02 &   777.75 &   603.07 &  6630.59 &   770.69 &   683.12 & \textbf{619.00} &   670.46 \\
neos-829552       & 10265.45 &  2906.35 &  6747.03 &  9010.09 &  3760.47 &  2687.66 & \textbf{1685.03} &  4602.34 \\
neos-839859       &   337.65 &   384.93 &   277.96 &  7120.80 &   173.49 &   439.95 & \textbf{428.35} &   295.92 \\
neos-892255       &220186.70 &187674.12 &267356.75 &257304.52 & 213861.48 &187294.79 &\textbf{164249.46} &173672.63 \\
neos12            & 44087.85 & 53639.77 & 87728.51 & 92286.19 & 11128.05 & 42895.16 &\textbf{19790.43} & 15487.38 \\
neos13            &  1368.44 &  1443.07 & 85810.06 & 84649.32 &  1444.75 &  1526.50 & 38838.92 &  1604.58 \\
neos18            &  2089.19 &  1899.67 & 12441.87 & 18554.88 &   546.53 &  2162.18 & \textbf{2107.20} &  1359.88 \\
ns1830653         &  5540.18 & 3964.29 & 10834.00  & 37564.13 &  3931.31 &  4706.58 & \textbf{3938.23} &  4473.86 \\
pigeon-10         & 36002.07 & 36002.36 & 36002.12 & 36003.96 & 36002.22 & 36003.55 & \textbf{36001.56} & 36002.18 \\
rail507           &   952.20 &   912.05 &  1174.57 &  4718.87 &   989.64 &   899.03 & \textbf{821.88} &  1043.87 \\
roll3000          &   321.50 &   536.84 &   341.09 &   929.54 &   147.57 &   261.69 & \textbf{ 69.01} &   311.69 \\
rout              &  1135.77 &   821.40 &   817.41 &  4624.89 &   132.84 &  3203.34 & \textbf{3029.28} &   846.96 \\
seymour1          &    95.41 &    95.40 &   131.77 &  2120.43 &    95.94 &   100.11 & \textbf{ 92.13} &    96.09 \\
\bottomrule
\end{tabular}
\end{table*}

\end{document}